\documentclass{article}

\PassOptionsToPackage{numbers, compress}{natbib}

\usepackage[preprint]{neurips_2026}

\usepackage[utf8]{inputenc} 
\usepackage[T1]{fontenc}    
\usepackage{hyperref}       
\usepackage{url}            
\usepackage{booktabs}       
\usepackage{amsfonts}       
\usepackage{nicefrac}       
\usepackage{microtype}      
\usepackage{xcolor}         
\usepackage{graphicx}
\usepackage{amsmath}

\usepackage{subcaption}
\usepackage{multirow}
\usepackage[table]{xcolor}

\title{XiYOLO: Energy-Aware Object Detection \\ via Iterative Architecture Search and Scaling}

\author{%
  Tony Tran \\
  Department of Research Computing\\
  University of Houston\\
  Houston, TX 77479 \\
  \texttt{thtran37@cougarnet.uh.edu} \\
  \And
  Richie R.~Suganda \\
  Department of Electrical and Computer Engineering \\
  University of Houston \\
  Houston, TX 77479 \\
  \texttt{rrsugand@cougarnet.uh.edu} \\
  \AND
  Bin Hu \\
  Department of Engineering Technology \\
  University of Houston \\
  Houston, TX 77479 \\
  \texttt{bhu11@central.uh.edu} \\
}

\begin{document}

\maketitle

\begin{abstract}
Object detection on heterogeneous edge devices must satisfy strict energy, latency, and memory constraints while still providing reliable perception for downstream autonomy. Existing energy-aware NAS methods often target limited deployment settings, while real energy remains difficult to optimize because it is highly device-dependent and costly to measure. We address these challenges with an energy-adaptive framework that combines an energy-aware XiResOFA search space, a two-stage energy estimator, and iterative search to identify a single energy-efficient base architecture. We then apply compound scaling to transform this base design into the \textbf{XiYOLO} family across deployment budgets, enabling interpretable accuracy--energy tradeoffs under sparse hardware measurements. Experiments on PascalVOC, COCO, and real-device deployment show that XiYOLO achieves a stronger energy--accuracy tradeoff than YOLO baselines. On PascalVOC, the medium XiYOLO model reaches \textbf{86.15 mAP50} while reducing energy relative to YOLOv12m by \textbf{20.6\%} on GPU and \textbf{35.9\%} on NPU. On COCO, XiYOLO reduces energy relative to YOLOv12 by up to \textbf{53.7\%} on GPU and \textbf{51.6\%} on NPU at the small scale. The proposed two-stage estimator also improves sample efficiency over a joint predictor under few-shot adaptation with only \textbf{2--20} target-device samples.
\end{abstract}

\section{Introduction}

Object detection is increasingly important for autonomous systems deployed on heterogeneous edge devices, including UAVs, UGVs, mobile robots, and embedded cameras. These systems support tasks such as search and rescue, infrastructure inspection, and remote monitoring, where perception must run directly on-device under tight latency, memory, compute, and energy constraints \cite{lin_mcunet_2020,burrello_dory_2021,banbury_micronets_2021,moss_ultra-low_2023}. Among these constraints, energy is especially important because it directly affects system endurance. Larger detectors often improve accuracy, but they also increase energy and latency, whereas smaller models are more efficient but may sacrifice robustness and small-object sensitivity \cite{tan_efficientdet_2020,xiong_mobiledets_2021,tu_femtodet_2023,alqahtani_benchmarking_2025}. As a result, the best detector is rarely universal across platforms or operating conditions, and edge perception should be treated as a \emph{budget-adaptive} problem.

Neural architecture search (NAS) offers a natural framework for discovering detector operating points along this accuracy--energy frontier \cite{cai_proxylessnas_2018,tan_mnasnet_2019,chen_detnas_2019,ghiasi_nas-fpn_2019,wang_nas-fcos_2020,sakuma_detofa_2023}, but applying NAS to energy-adaptive edge perception remains difficult. Many existing methods still emphasize a single final model or a limited set of hardware settings rather than scalable detector families across heterogeneous devices \cite{xiong_mobiledets_2021,gupta_efficient_2024,chen_hardware-aware_2025,tran_elastic_2026}. More importantly, energy is hard to optimize directly: FLOPs, MACs, and parameter count often fail to predict measured energy because energy also depends on memory movement, operator composition, and platform-specific execution behavior \cite{cai_neuralpower_2017,guo_estimating_2025,zhang_ampere_2025}. The central challenge is therefore not only how to search for efficient detectors, but how to make energy-aware detector design transferable across heterogeneous platforms when accurate hardware energy measurements are scarce.

In this paper, we address three key challenges in energy-adaptive object detection for heterogeneous edge devices: designing a search space with meaningful accuracy--energy tradeoffs, searching that space efficiently, and enabling energy-aware search when hardware energy data is scarce. To address these challenges, we design a novel XiResOFA search block, use an iterative search strategy over the backbone, FPN, and PAN, and develop a two-stage energy approximation model with a generic predictor and a hardware-specific residual. Together, these components identify a single energy-efficient base architecture, which we scale into a family of deployable detectors spanning different accuracy--energy operating regimes. In this way, the proposed framework treats edge detector design not as the search for a single efficient model, but as the search for budget-adaptive operating points aligned with heterogeneous hardware constraints.

Our contributions are summarized as follows:
\begin{itemize}
    \item We formulate energy-adaptive perception as the design of an energy-efficient base detector architecture that can be scaled across deployment budgets on heterogeneous edge devices, and we instantiate this idea with XiYOLO, whose medium PascalVOC model achieves the highest \textbf{86.15 mAP50} among all compared medium-sized detectors.
    \item We propose an iterative neural architecture search framework for object detection that progressively searches the backbone, FPN, and PAN to identify an energy-efficient base architecture. After scaling, XiYOLO achieves strong accuracy--energy tradeoffs on both PascalVOC and COCO: on PascalVOC, the medium model improves over YOLOv12m by \textbf{+0.07 mAP50} while reducing energy by \textbf{20.6\%} on GPU and \textbf{35.9\%} on NPU; on COCO, the medium model reduces energy relative to YOLOv12m by \textbf{24.5\%} on GPU and \textbf{38.8\%} on NPU while maintaining competitive accuracy.
    \item We introduce an energy-aware search space based on a novel XiResOFA block with controllable compression ratio, kernel size, and lite/full attention choices, enabling detector search over interpretable accuracy--energy tradeoffs. Randomly sampled XiResOFA medium models reduce average energy by \textbf{21.1\%} relative to the YOLOv12 medium baseline, while the iterative search space yields about \textbf{3.5\%} higher average predicted mAP50 than the global search space.
    \item We develop a two-stage energy approximation model that combines a generic architecture--device energy prior with a lightweight hardware-specific residual model, reducing the amount of measured energy data required for energy-aware search, and we validate its sample efficiency on HW-NAS-Bench under few-shot adaptation with only \textbf{2--20} target-device samples.
\end{itemize}

\section{Related Work}

\paragraph{Efficient object detection for onboard deployment.}
Modern object detection has evolved from early real-time one-stage detectors such as SSD and YOLO to stronger multiscale architectures such as FPN, FCOS, and EfficientDet \cite{liu_ssd_2016,redmon_you_2016,lin_feature_2017,tian_fcos_2022,tan_efficientdet_2020}. For resource-constrained deployment, prior work has developed lightweight detector families including MobileDets, TinyissimoYOLO, and FemtoDet, highlighting the need to balance detection accuracy against runtime and model cost on edge hardware \cite{xiong_mobiledets_2021,moosmann_tinyissimoyolo_2023,tu_femtodet_2023}. In aerial and onboard settings, recent efforts have further emphasized efficient detection under limited compute and energy budgets \cite{zhou_swdet_2023,liew_altitude-informed_2025,nghiem_leaf-yolo_2025,geng_efficient_2025}. These works establish the importance of efficient onboard perception, but they typically target a single deployed detector rather than a set of models spanning different energy budgets.

\paragraph{Hardware-aware neural architecture search for detection.}
Neural architecture search has become an effective tool for automating efficient model design, and hardware-aware variants incorporate deployment-dependent costs such as latency or memory into the search objective \cite{cai_proxylessnas_2018,tan_mnasnet_2019,wu_fbnet_2019}. In object detection, prior work has searched detector backbones, feature pyramids, and full detector pipelines, including DetNAS, NAS-FPN, NAS-FCOS, MobileDets, and DetOFA \cite{chen_detnas_2019,ghiasi_nas-fpn_2019,wang_nas-fcos_2020,xiong_mobiledets_2021,sakuma_detofa_2023}. More recent work has moved toward edge- and hardware-aware detector NAS \cite{gupta_efficient_2024,chen_hardware-aware_2025,tran_elastic_2026}. However, these approaches generally optimize for a single final architecture or a single deployment target. In contrast, our setting requires a \emph{budget-adaptive detector family} that spans multiple operating points along the accuracy--energy frontier for onboard use.

\paragraph{Energy modeling for neural networks.}
Practical energy-aware search depends on being able to estimate hardware energy efficiently. Early work such as NeuralPower showed that neural network energy can be predicted from architectural characteristics, while more recent approaches have incorporated both network structure and device information for improved modeling \cite{cai_neuralpower_2017,guo_estimating_2025,zhang_ampere_2025}. At the same time, these studies reinforce that energy cannot be reliably reduced to FLOPs, MACs, or parameter count alone, since measured energy is strongly influenced by memory movement, operator composition, and device-specific execution behavior \cite{cai_neuralpower_2017,guo_estimating_2025}. This limitation is particularly important for detector NAS, where dense hardware measurements across a large search space are costly to obtain. Our work builds on this observation by coupling detector search with a two-stage energy approximation strategy designed for sparse hardware measurements.

\begin{figure}[t]
    \centering
    \includegraphics[width=\linewidth]{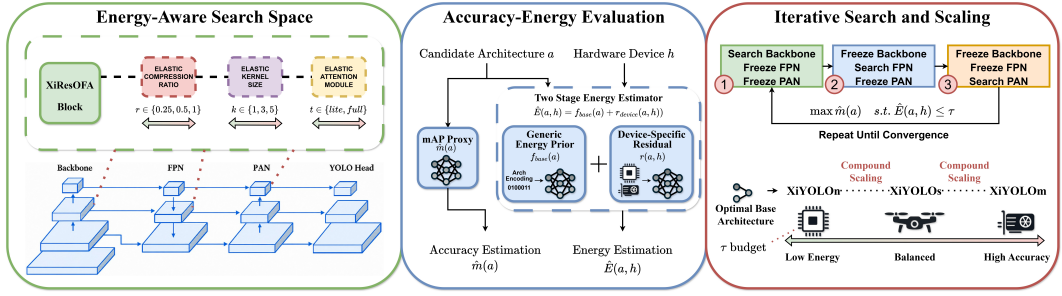}
    \caption{\textbf{Energy-aware NAS framework.} Candidate detectors from a searchable energy-aware architecture space are ranked by an mAP proxy and a two-stage energy estimator, then refined iteratively to obtain scalable models under hardware-specific energy constraints.}
    \label{fig:main}
\end{figure}

\section{Methodology}

\subsection{Problem Formulation and Framework Overview}

We address onboard object detection under realistic robotic deployment constraints, where a perception model must operate within limited energy, latency, and memory budgets on a target hardware platform \cite{lin_mcunet_2020,burrello_dory_2021,banbury_micronets_2021,moss_ultra-low_2023}. Let $h$ denote the target hardware platform and let $a \in \mathcal{A}$ denote a candidate detector architecture drawn from search space $\mathcal{A}$. We denote by $\hat{m}(a)$ the predicted detection accuracy of architecture $a$, measured through an mAP proxy, and by $\hat{E}(a,h)$ its estimated energy on hardware $h$. Our goal is to identify a single energy-efficient base detector that maximizes detection quality while satisfying a target energy budget. We therefore formulate architecture search as
\begin{equation*}
a^\star = \arg\max_{a \in \mathcal{A}} \hat{m}(a)
\quad \text{s.t.} \quad
\hat{E}(a,h) \le \tau ,
\label{eq:main_objective}
\end{equation*}
where $\tau$ is the deployment energy budget for hardware platform $h$.

Rather than directly searching for multiple detectors, the framework first identifies a single optimized base architecture and then derives deployment variants by scaling that architecture to different budget levels. This separates architecture discovery from deployment adaptation: iterative search finds the base design, while post-search scaling produces high-accuracy, balanced, and low-energy variants \cite{cai_once_2020,sakuma_detofa_2023,tran_elastic_2026}. As shown in Figure~\ref{fig:main}, candidate detectors are sampled from an energy-aware search space over the backbone, FPN, and PAN, then evaluated using a detection-quality proxy and a two-stage energy estimator that combines a generic architecture predictor with a lightweight device-specific correction term \cite{cai_neuralpower_2017,guo_estimating_2025,zhang_ampere_2025}. These signals guide an iterative NAS loop toward a single energy-efficient operating point, after which the searched base architecture is scaled to match different resource budgets \cite{cai_once_2020}.

\begin{figure}[t]
    \centering
    \begin{subfigure}[b]{0.39\textwidth}
        \centering
        \includegraphics[width=\textwidth]{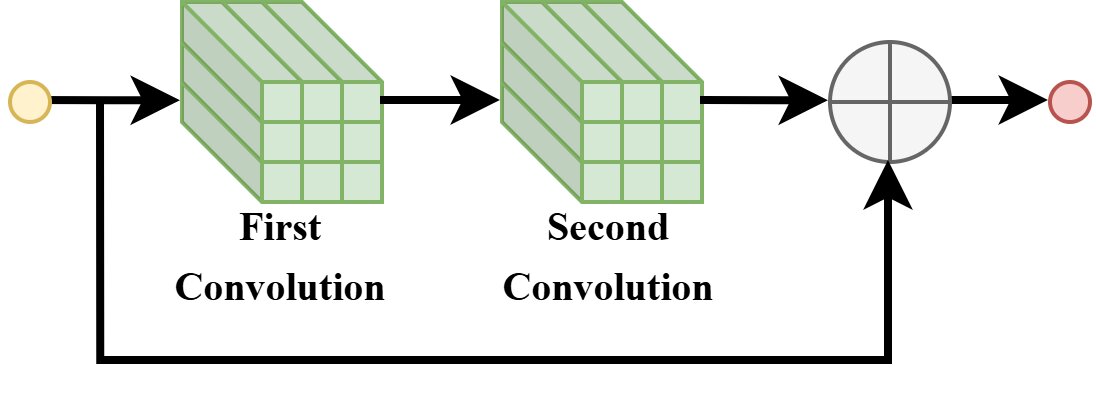}
        \caption{Default Bottleneck Block}
        \label{fig:a}
    \end{subfigure}
    \hfill
    \begin{subfigure}[b]{0.58\textwidth}
        \centering
        \includegraphics[width=\textwidth]{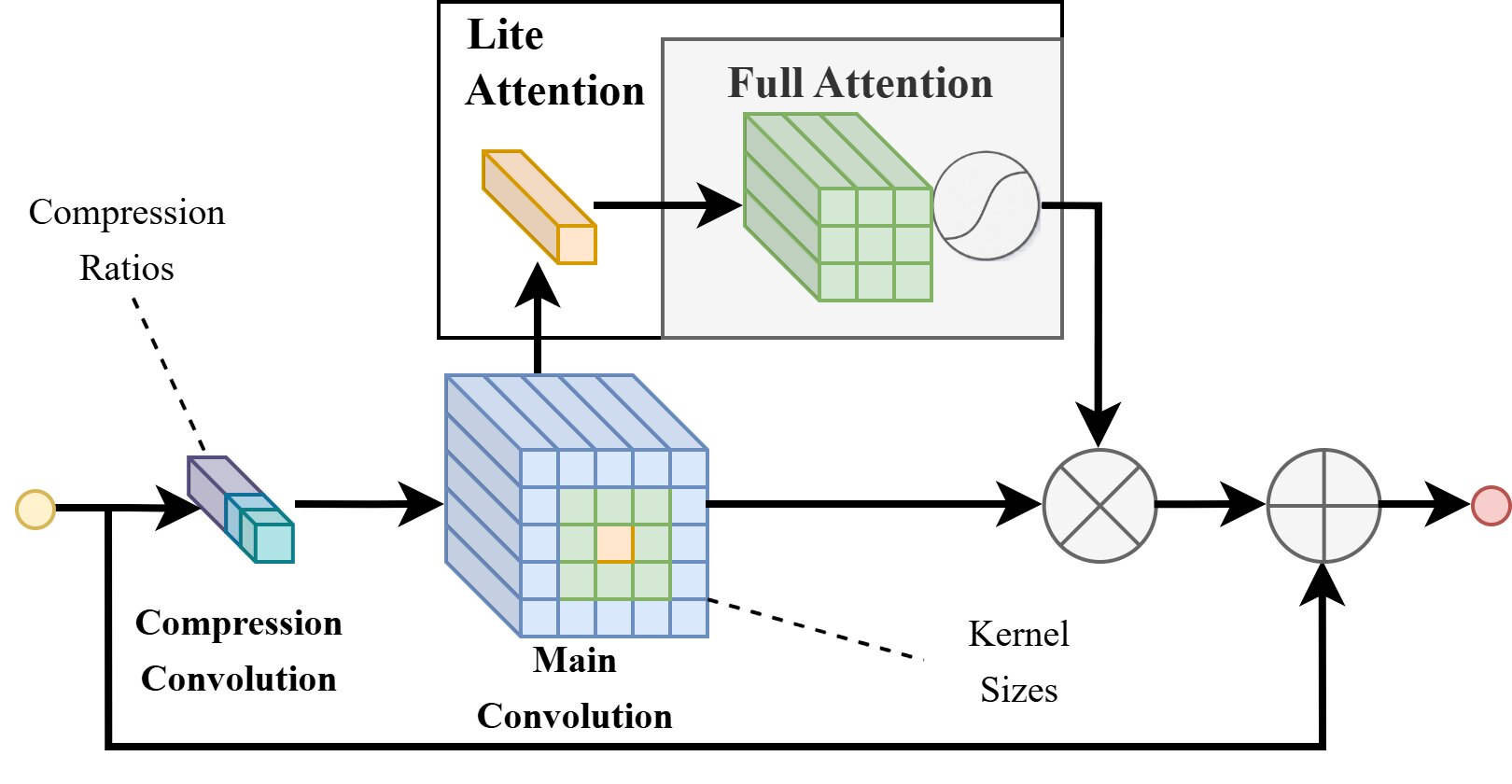}
        \caption{XiResOFA Block}
        \label{fig:b}
    \end{subfigure}
    \caption{Comparison between (a) the standard bottleneck block~\cite{he_deep_2016} and (b) the proposed XiResOFA block. XiResOFA augments a residual design with elastic architectural choices for energy-aware detector search.}
    \label{fig:blocks}
\end{figure}

\subsection{Energy-Aware Search Space}

A central design requirement of the search space is that its architectural choices should have a direct and interpretable relationship to both detection quality and hardware energy. Figure~\ref{fig:blocks} compares the standard residual bottleneck block with the proposed \emph{XiResOFA} block. The standard bottleneck provides stable optimization through identity shortcuts \cite{he_deep_2016}, but it does not directly expose elastic operator-level choices suitable for energy-aware detector search. In contrast, XiResOFA preserves a residual shortcut while introducing configurable architectural choices compatible with once-for-all search \cite{cai_once_2020}. This design makes the block both trainable and searchable, while allowing its internal structure to adapt to different accuracy--energy operating points.

\begin{figure}[t]
    \centering
    \begin{subfigure}[b]{0.39\textwidth}
        \centering
        \includegraphics[width=\textwidth]{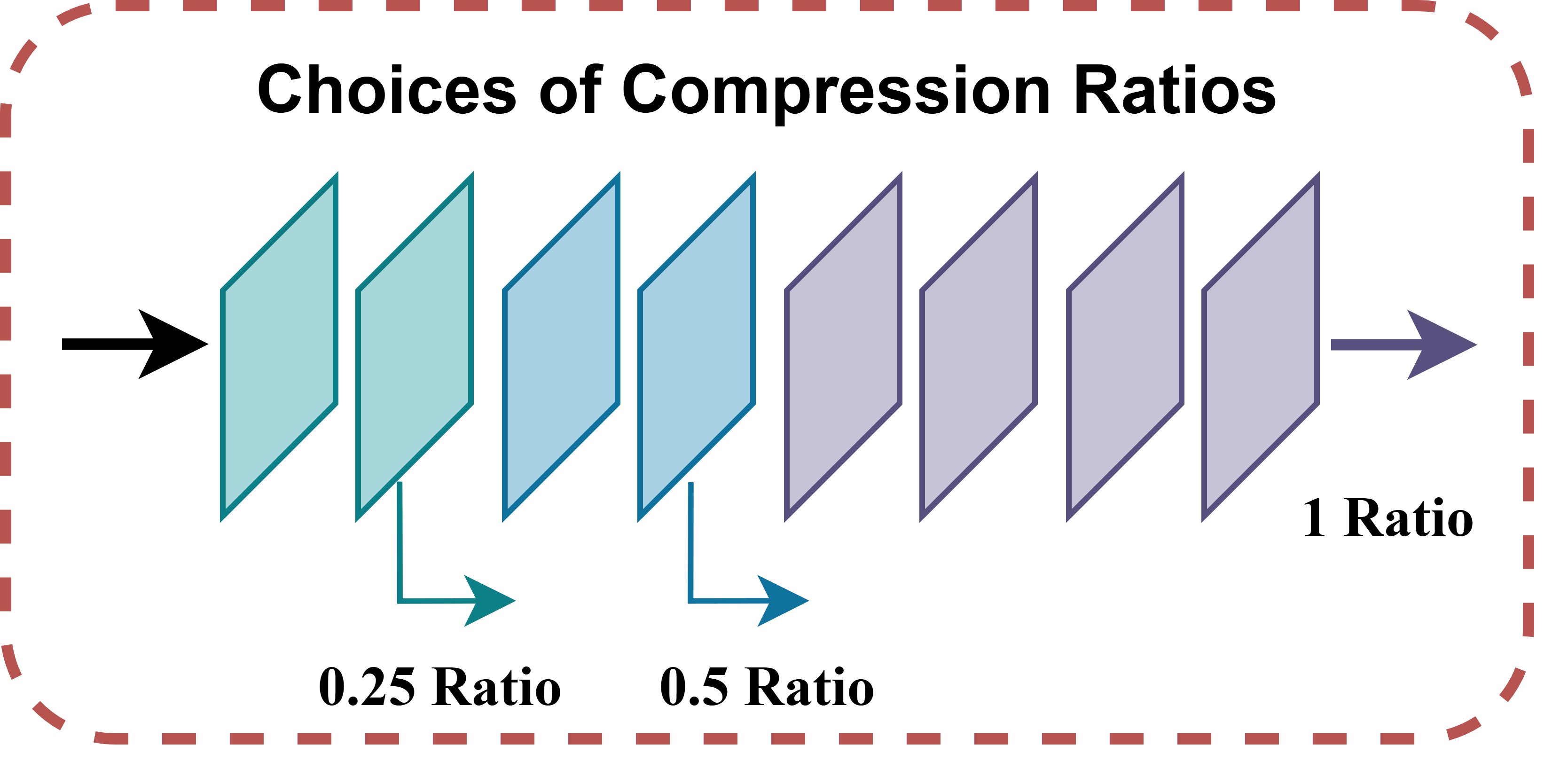}
        \caption{Compression Ratio}
        \label{fig:choice1}
    \end{subfigure}
    \hfill
    \begin{subfigure}[b]{0.30\textwidth}
        \centering
        \includegraphics[width=\textwidth]{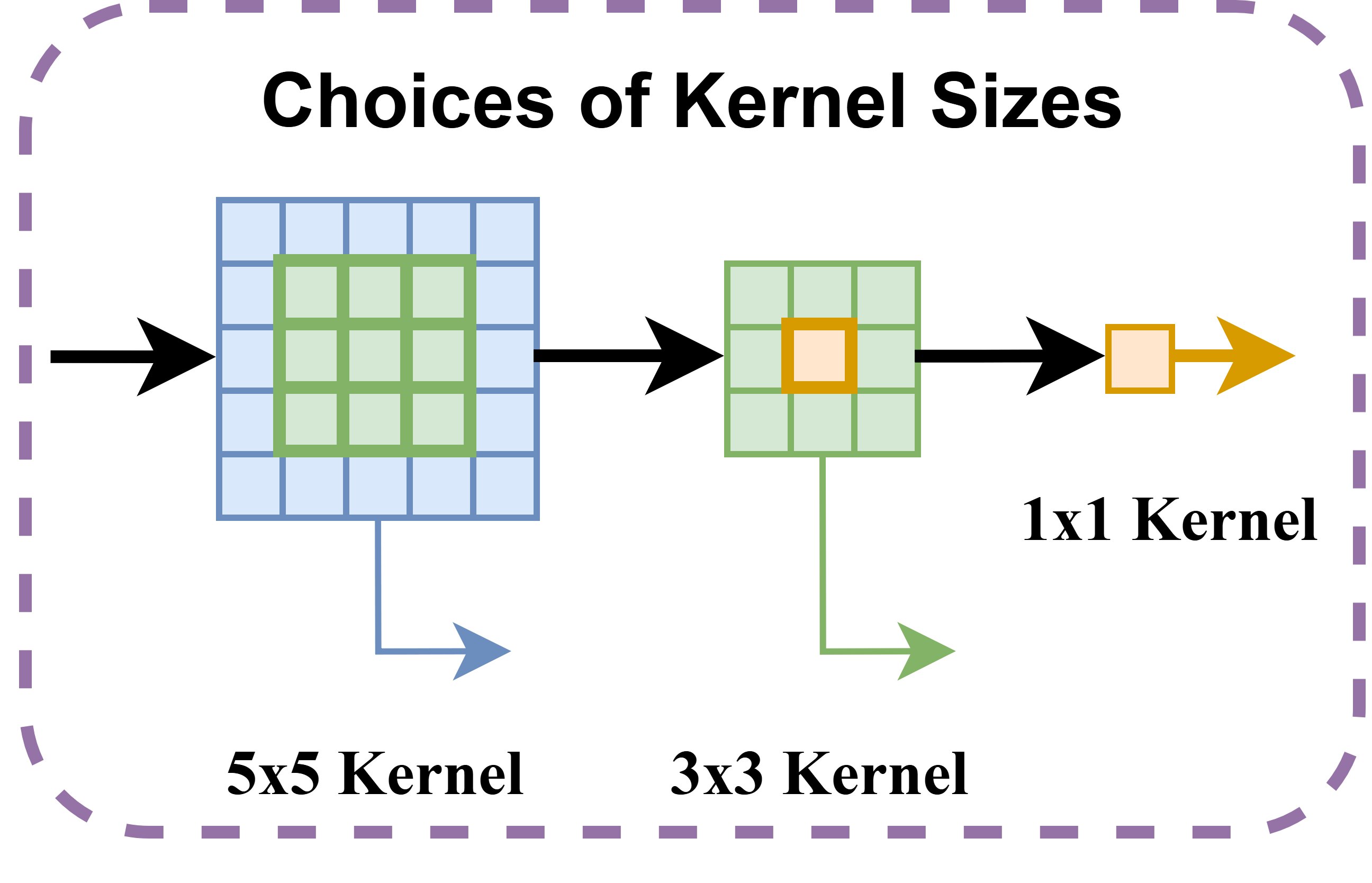}
        \caption{Kernel Size}
        \label{fig:choice2}
    \end{subfigure}
    \hfill
    \begin{subfigure}[b]{0.26\textwidth}
        \centering
        \includegraphics[width=\textwidth]{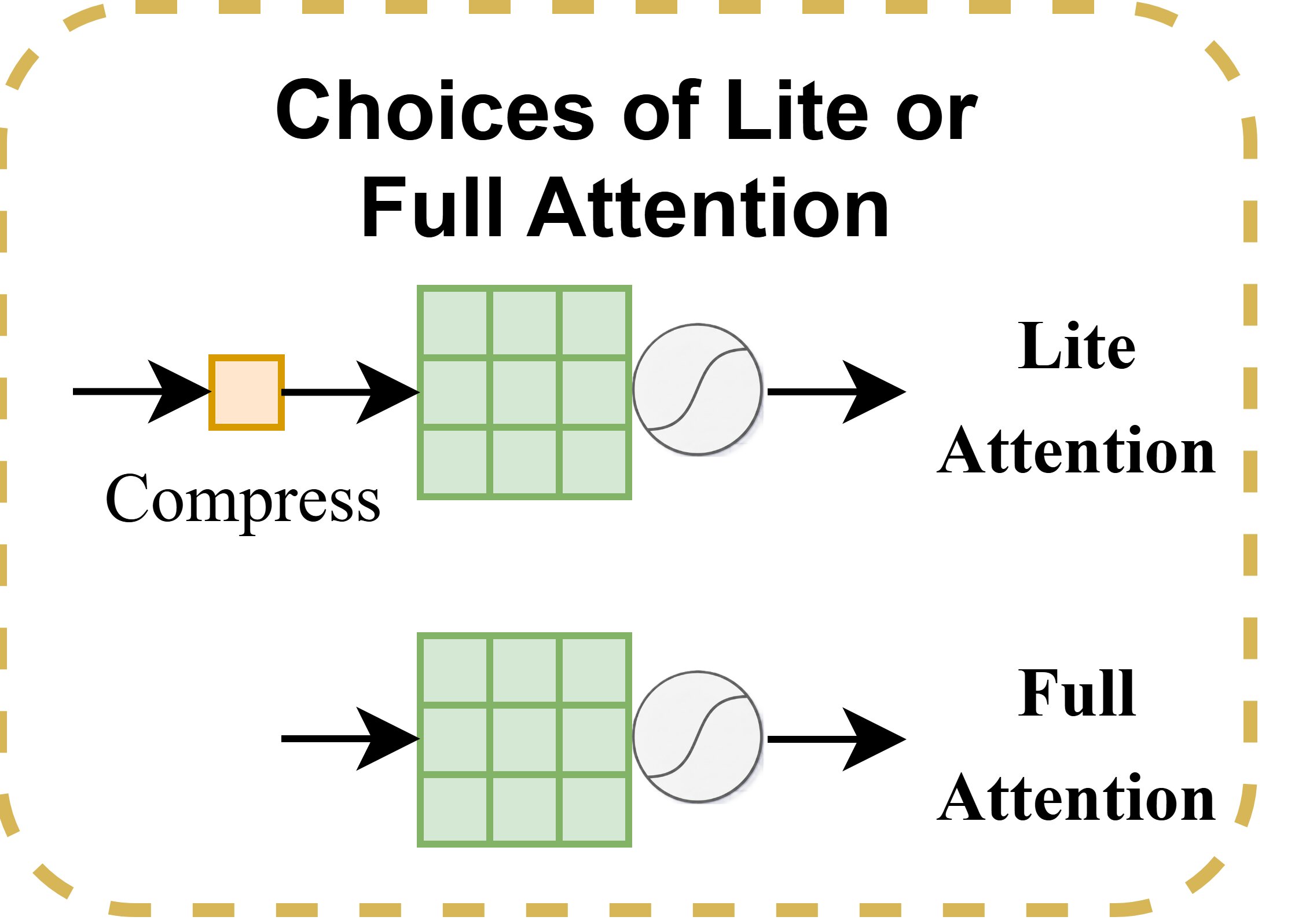}
        \caption{Lite/Full Attention}
        \label{fig:choice3}
    \end{subfigure}
    \caption{Elastic architectural choices in XiResOFA: selectable (a) compression ratios, (b) kernel sizes, and (c) lite or full attention mechanisms.}
    \label{fig:choices}
\end{figure}

Figure~\ref{fig:choices} illustrates the three elastic choices exposed by each XiResOFA block. Each block is parameterized by a tuple $b = (r,k,t)$, where $r \in \{1,0.5,0.25\}$ is the compression ratio, $k \in \{1,3,5\}$ is the kernel size, and $t \in \{\text{lite}, \text{full}\}$ is the attention type. These dimensions induce clear accuracy--energy tradeoffs. Lower compression preserves more channels and typically improves representational strength, but increases arithmetic cost and activation movement. Larger kernels expand the receptive field and may improve contextual reasoning, especially for multiscale detection, but also increase energy consumption. Full attention provides stronger feature recalibration than a lightweight alternative, but introduces additional compute and memory overhead. By embedding these choices within a shared supernet, the search space can represent a range of candidate architectures from aggressively efficient configurations to higher-capacity detectors \cite{cai_once_2020,tan_mnasnet_2019,wu_fbnet_2019}.

We apply XiResOFA consistently across the major stages of a one-stage detector, including the backbone, FPN, and PAN. The backbone extracts hierarchical feature representations, the FPN performs multiscale semantic fusion, and the PAN strengthens bottom-up aggregation for localization and small-object reasoning \cite{lin_feature_2017,liu_path_2018}. A candidate detector is therefore written as $a = \big(a^{\text{back}}, a^{\text{fpn}}, a^{\text{pan}}\big)$, where each component denotes a sequence of XiResOFA block choices. The full search space is $\mathcal{A} = \mathcal{A}_{\text{back}}\times\mathcal{A}_{\text{fpn}}\times\mathcal{A}_{\text{pan}}$. Using the same searchable block family across all three stages provides a coherent detector-level search space rather than imposing unrelated search rules across different modules. It also enables efficient weight sharing across subnetworks, which substantially reduces the cost of exploring detector candidates during search \cite{cai_once_2020,sakuma_detofa_2023}.

\begin{figure}[t]
    \centering
    \includegraphics[width=\linewidth]{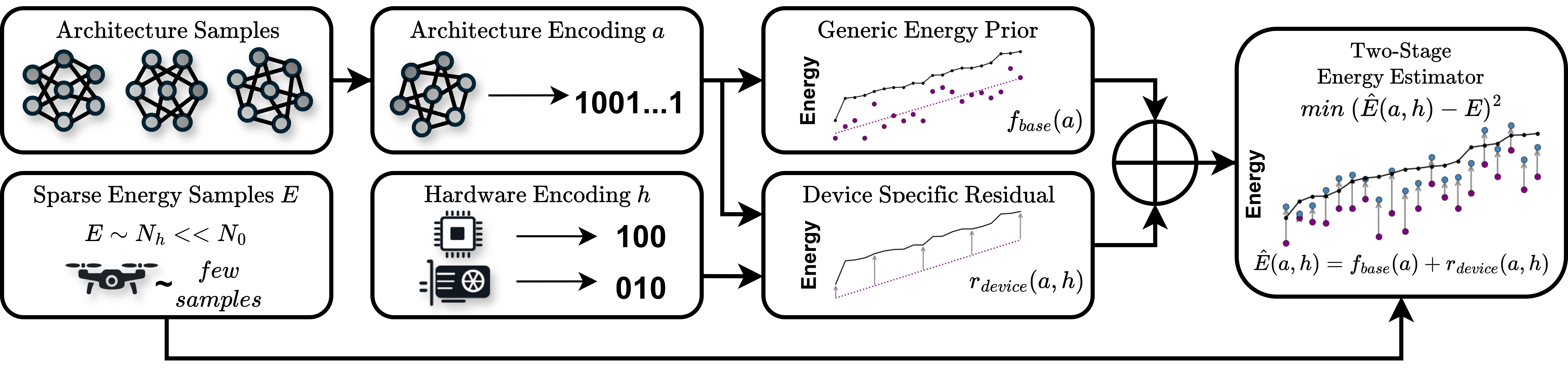}
    \caption{Overview of the proposed two-stage energy estimator. Sparse energy samples from the target hardware are combined with architecture and hardware encodings to learn a generic energy prior and a device-specific residual.}
    \label{fig:energyf}
\end{figure}

\subsection{Two-Stage Energy Estimation}

A major obstacle in energy-aware detector search is that true hardware energy is expensive to measure at scale \cite{cai_neuralpower_2017,guo_estimating_2025,zhang_ampere_2025}. Simple analytical surrogates such as FLOPs, MACs, or parameter count are often insufficient because measured energy also depends on memory traffic, intermediate activations, operator composition, and hardware-specific execution behavior \cite{cai_neuralpower_2017,guo_estimating_2025}. For object detectors, this makes direct energy optimization particularly difficult, since exhaustive profiling over a large search space is costly and often impractical.

Figure~\ref{fig:energyf} illustrates the proposed two-stage energy estimation pipeline. Let $N_0$ denote the full number of candidate architectures in the search space and let $N_h$ denote the number of measured energy samples available on target hardware $h$, where in practice $N_h \ll N_0$.
Given architecture samples and sparse energy measurements from the target device, we encode both the detector architecture and the hardware target into compact representations and decompose energy prediction into a generic prior and a device-specific correction. For a candidate architecture $a$ on hardware $h$, we estimate energy as
\begin{equation*}
\hat{E}(a,h) = f_{\text{base}}(a) + r(a,h),
\label{eq:energy_model}
\end{equation*}
where $f_{\text{base}}(a)$ is a generic pretrained energy predictor and $r(a,h)$ is a lightweight hardware-specific residual model. The first stage captures transferable architecture-level energy structure, while the second stage adapts that prior to device-specific effects such as backend implementation and memory hierarchy \cite{guo_estimating_2025,zhang_ampere_2025}. Because the residual only models the remaining hardware-specific discrepancy, it can be calibrated with relatively few measured samples, that is, using only $N_h$ target-device measurements instead of exhaustively profiling all $N_0$ candidates. During search, this two-stage predictor replaces direct hardware measurement and enables practical hardware-aware ranking of detector candidates under sparse energy supervision.

\subsection{Energy-Aware Iterative Search and Scaling}

Jointly searching over the backbone, FPN, and PAN creates a large combinatorial design space and can make hardware-aware detector NAS expensive and difficult to optimize \cite{chen_detnas_2019,ghiasi_nas-fpn_2019,wang_nas-fcos_2020,sakuma_detofa_2023}. These components also influence accuracy and energy differently: backbone choices largely determine representational strength and compute load, while FPN and PAN choices affect multiscale fusion, feature reuse, and memory traffic \cite{lin_feature_2017,liu_path_2018}. To exploit this structure, we use an iterative search procedure that progressively refines detector components rather than performing a single monolithic joint search.

At iteration $t$, let the current detector be $a_t = \big(a_t^{\text{back}}, a_t^{\text{fpn}}, a_t^{\text{pan}}\big)$. We alternately optimize each detector stage while conditioning on the current choices of the others:
\begin{align*}
a_{t+1}^{\text{back}}
&=
\arg\max_{b \in \mathcal{A}_{\text{back}}}
\hat{m}\!\left(b, a_t^{\text{fpn}}, a_t^{\text{pan}}\right)
\quad
\text{s.t.}
\quad
\hat{E}\!\left(b, a_t^{\text{fpn}}, a_t^{\text{pan}}, h\right) \le \tau,
\\
a_{t+1}^{\text{fpn}}
&=
\arg\max_{f \in \mathcal{A}_{\text{fpn}}}
\hat{m}\!\left(a_{t+1}^{\text{back}}, f, a_t^{\text{pan}}\right)
\quad
\text{s.t.}
\quad
\hat{E}\!\left(a_{t+1}^{\text{back}}, f, a_t^{\text{pan}}, h\right) \le \tau,
\\
a_{t+1}^{\text{pan}}
&=
\arg\max_{p \in \mathcal{A}_{\text{pan}}}
\hat{m}\!\left(a_{t+1}^{\text{back}}, a_{t+1}^{\text{fpn}}, p\right)
\quad
\text{s.t.}
\quad
\hat{E}\!\left(a_{t+1}^{\text{back}}, a_{t+1}^{\text{fpn}}, p, h\right) \le \tau.
\end{align*}
Each step selects the highest-scoring candidate that satisfies the energy budget, allowing the backbone and neck to co-adapt over successive iterations. This stage-wise refinement reduces search complexity while preserving the main accuracy--energy tradeoff. After $T$ iterations, the final searched base architecture is $a^\star = a_T$.

After search, we derive a detector family by scaling the selected base architecture to different deployment budgets. Let $S(a^\star,\alpha)$ denote a scaling operator with scale factor $\alpha$. We obtain the final deployment variants as
\begin{equation*}
a^{(\ell)} = S(a^\star,\alpha_\ell),
\qquad
\ell \in \{n,s,m\},
\label{eq:scaling_rule}
\end{equation*}
where $n$, $s$, and $m$ denote nano, small, and medium variants. The searched model serves as the balanced reference design, while compound scaling adjusts width, depth, or related stage capacity to produce higher- and lower-budget variants \cite{cai_once_2020}. In this way, iterative NAS identifies the core detector design, and post-search scaling adapts it to different operating regimes without requiring separate searches for each deployment point.

\section{Experiments}

\subsection{Experimental Setup}
\label{sec:setup}

\paragraph{Search space.}
We construct the detector search space by starting from the base YOLOv12n architecture \cite{tian_yolov12_2025} and replacing each bottleneck block with the proposed XiResidualOFA block. Each XiResidualOFA block exposes three choices: compression ratio, kernel size, and attention type. For compression, we use ratios of $\{0.25, 0.5, 1\}$, where smaller ratios preserve more channels and generally improve accuracy at higher energy cost, while larger ratios are more efficient but less expressive. For kernel size, we use $\{1,3,5\}$, where larger kernels provide stronger spatial context at higher energy cost. For attention, we consider \emph{full} and \emph{lite} variants, where full attention is more expressive but more expensive. The resulting detector search space is partitioned into a backbone space with $18^{2}$ configurations, an FPN space with $18^{4}$ configurations, and a PAN space with $18^{4}$ configurations.

\paragraph{Training protocol.}
On PascalVOC \cite{everingham_pascal_2010}, we train the supernet for 300 epochs on one RTX 4090 GPU with batch size 64 using the largest supernet configuration, followed by 300 epochs of uniform subnet training by randomly sampling one subnet per batch. We then sample 1000 subnetworks to train an mAP proxy and an energy proxy, both implemented as MLPs with hidden size 400, and perform iterative search for 4 iterations over the backbone, FPN, and PAN to obtain a single optimized base architecture. The searched subnet is fine-tuned for 300 epochs, then compound scaled to small and medium variants, which are also fine-tuned for 300 epochs. On COCO, we follow the same procedure and search space, but train using 8 V100 GPUs. The searched architectures for PascalVOC and COCO are obtained independently. We refer to the resulting detector family as \textbf{XiYolo}.

\paragraph{Deployment platforms and baselines.}
We evaluate the searched models against YOLO baselines on the ModalAI Sentinel Development Drone, which contains a Qualcomm QRB5165 CPU, a Qualcomm Adreno 650 GPU, and a 15 TOPS NPU. We compare against YOLOv5 \cite{jocher_yolov5_2020}, YOLOv8 \cite{jocher_yolov}, YOLO11 \cite{jocher_yolov}, and YOLOv12 \cite{tian_yolov12_2025} at nano, small, and medium scales. All models are exported to FP16 TFLite, and we measure energy per inference, cumulative energy over time, latency, and detection accuracy. Power is monitored using the VOXL Power Module v3. We estimate inference energy by subtracting idle power from measured power to obtain active inference power, then multiplying by inference time and aggregating over 1000 samples.

\begin{figure}[t]
  \centering
  \includegraphics[width=\textwidth]{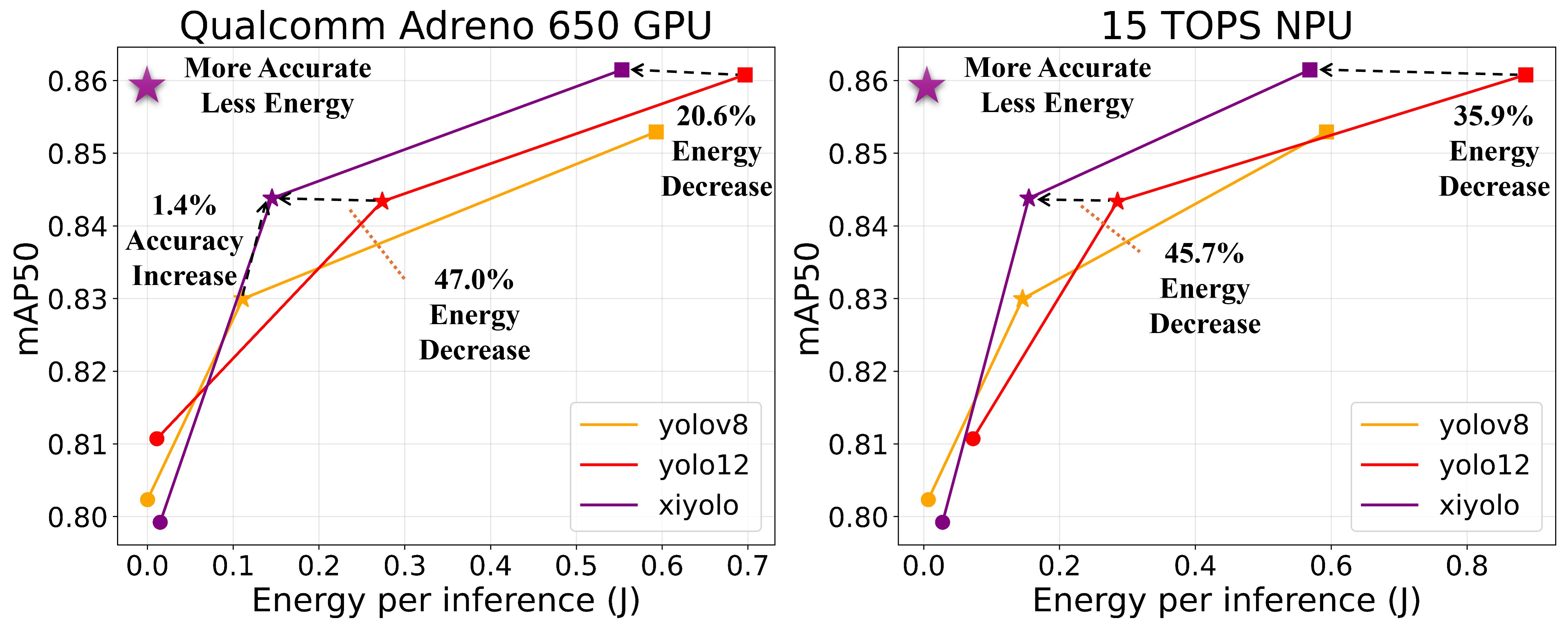}
  \caption{Comparisons with other YOLO-series~\cite{jocher_yolov, tian_yolov12_2025} on PascalVOC dataset in terms of energy and accuracy on the GPU (left) and NPU (right).}
  \label{fig:pareto}
\end{figure}

\begin{figure}[t]
  \centering
  \includegraphics[width=\textwidth]{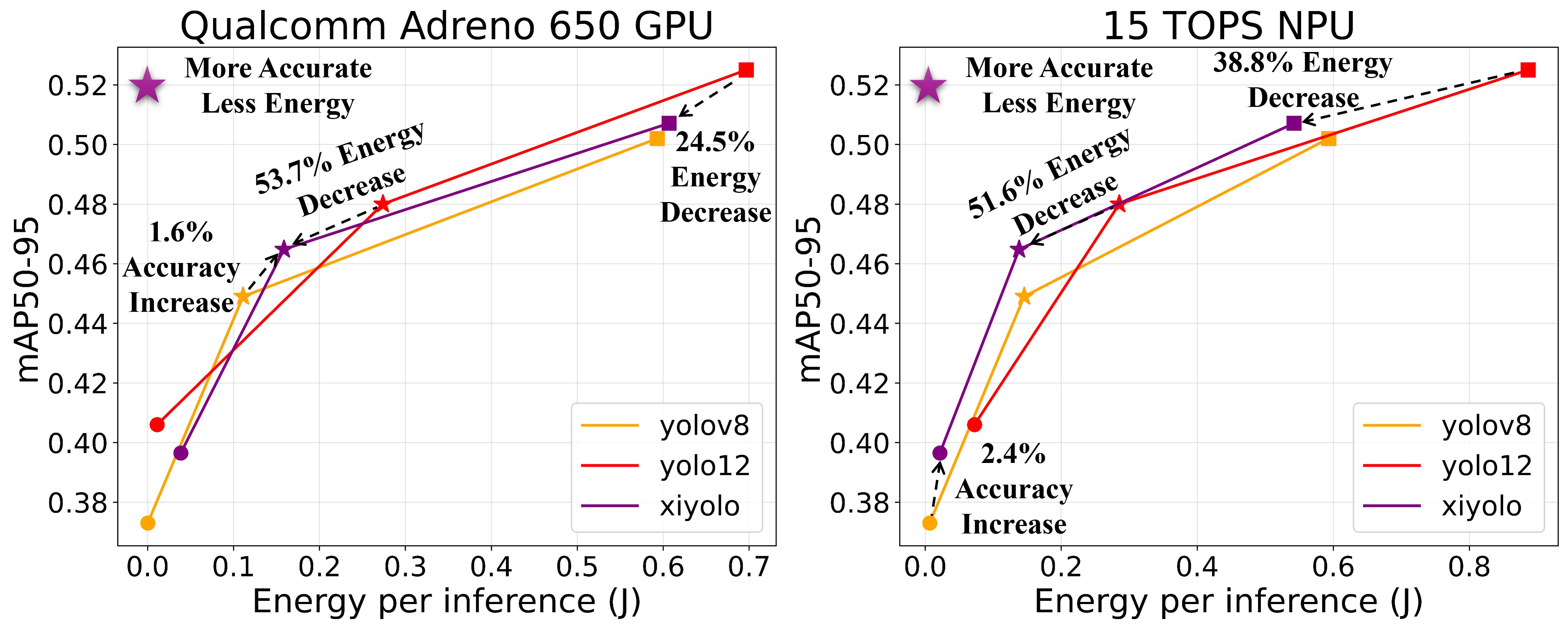}
  \caption{Comparisons with other YOLO-series~\cite{jocher_yolov, tian_yolov12_2025} on COCO dataset in terms of energy and accuracy on the GPU (left) and NPU (right).}
  \label{fig:paretococo}
\end{figure}

\subsection{Accuracy and Energy Tradeoffs}

We first evaluate the deployment quality of the searched models by comparing detection accuracy and energy jointly. Figures~\ref{fig:pareto} and~\ref{fig:paretococo} compare XiYOLO against YOLOv8 and YOLOv12 on PascalVOC and COCO, respectively, on the Qualcomm Adreno 650 GPU and the 15 TOPS NPU. Since each model is evaluated from a fixed trained checkpoint, accuracy remains constant across devices, while energy changes with the deployment backend. The key result is that XiYOLO consistently improves the accuracy--energy tradeoff relative to YOLOv12, with the clearest gains at the small and medium scales.

On PascalVOC, XiYOLO achieves the strongest overall tradeoff on both GPU and NPU. At the medium scale, XiYOLO reaches the highest \textbf{86.15 mAP50}, improving over YOLOv12m by \textbf{+0.07 mAP50}, while reducing energy by \textbf{20.6\%} on GPU and \textbf{35.9\%} on NPU. At the small scale, XiYOLO reaches \textbf{84.38 mAP50}, slightly above YOLOv12s at \textbf{84.34 mAP50}, while reducing energy by \textbf{47.0\%} on GPU and \textbf{45.7\%} on NPU. Relative to YOLOv8, XiYOLO also provides clear accuracy gains, including \textbf{+1.4\%} mAP50 at the small scale. These results show that on PascalVOC, XiYOLO improves accuracy while simultaneously lowering energy, especially in the practically relevant small and medium operating regimes.

On COCO, XiYOLO again shows a favorable accuracy--energy tradeoff, although the gains are more selective. At the medium scale, XiYOLO reduces energy relative to YOLOv12m by \textbf{24.5\%} on GPU and \textbf{38.8\%} on NPU, while maintaining competitive accuracy (\textbf{50.71} vs.\ \textbf{52.50 mAP50-95}). At the small scale, XiYOLO improves over YOLOv8s by \textbf{+1.6\%} mAP50-95 on GPU and over YOLOv12s by \textbf{+2.4\%} mAP50-95 on NPU, while also reducing energy relative to YOLOv12s by \textbf{53.7\%} on GPU and \textbf{51.6\%} on NPU. Overall, Figures~\ref{fig:pareto} and~\ref{fig:paretococo} show that XiYOLO consistently shifts the accuracy--energy frontier in a favorable direction, with especially strong gains against YOLOv12 and the most consistent benefits appearing at the small and medium scales.

\begin{figure}[t]
  \centering
  \includegraphics[width=\textwidth]{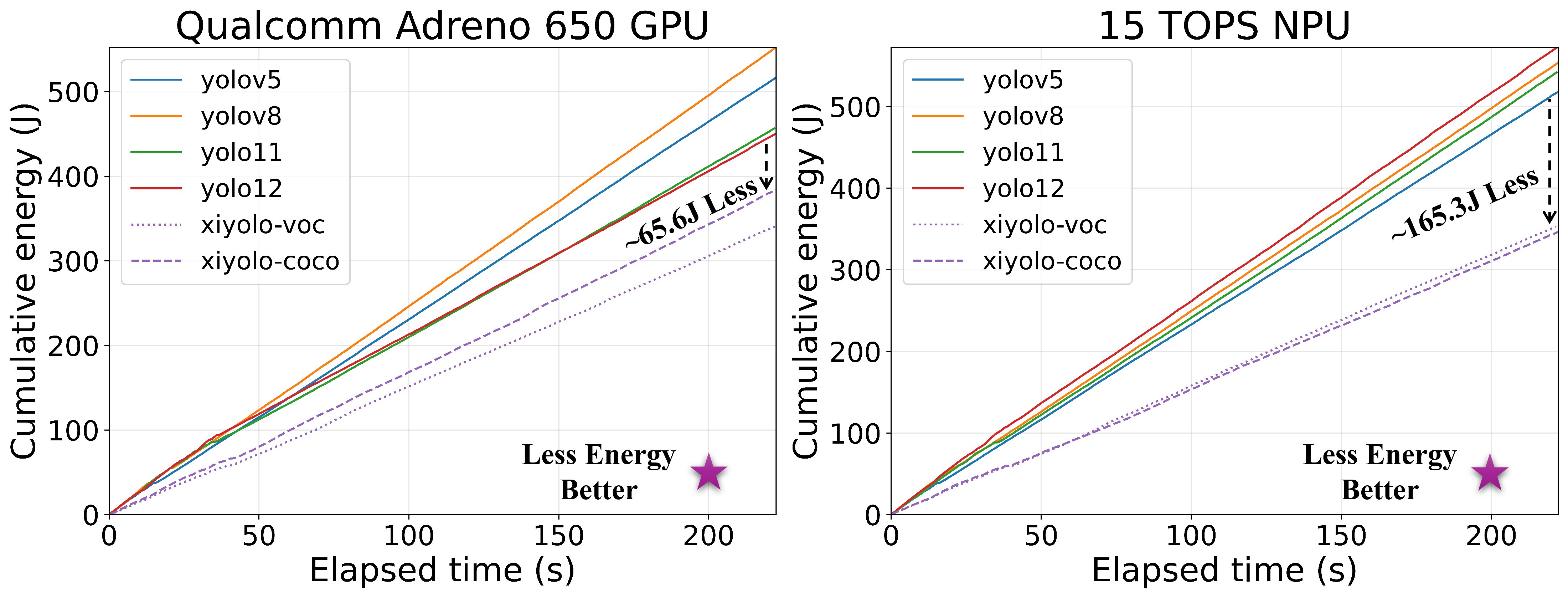}
  \caption{Energy consumption vs. time benchmarks against other YOLO-series~\cite{jocher_yolov5_2020, jocher_yolov, tian_yolov12_2025} on the GPU (left) and NPU (right) for the medium model scales.}
  \label{fig:energy}
\end{figure}

\subsection{Deployment Energy and Scaling Behavior}

We next examine how the searched architectures behave during sustained deployment. Figure~\ref{fig:energy} reports cumulative energy over time for medium-scale models on the Qualcomm Adreno 650 GPU and the 15 TOPS NPU. The clearest result is that both \textbf{XiYOLO-VOC} and \textbf{XiYOLO-COCO} accumulate substantially less energy over time than the YOLO baselines on both devices. This shows that the searched architectures preserve their efficiency advantage not only in per-inference measurements, but also under continuous deployment.

On the GPU, both XiYOLO variants remain below all YOLO baselines throughout the benchmark horizon, with \textbf{XiYOLO-VOC} achieving the lowest cumulative energy overall. By the end of the run, XiYOLO-COCO uses roughly \textbf{65.6 J} less cumulative energy than the strongest YOLO baseline. On the NPU, the separation is even larger: both XiYOLO variants again remain below all YOLO baselines, and XiYOLO-COCO achieves the lowest cumulative energy, using about \textbf{165.3 J} less cumulative energy than the strongest YOLO baseline. Overall, Figure~\ref{fig:energy} shows that the searched XiYOLO architectures deliver the best deployment-time energy behavior among all compared medium-sized models on both accelerator backends.

\begin{figure}[t]
  \centering
  \includegraphics[width=\textwidth]{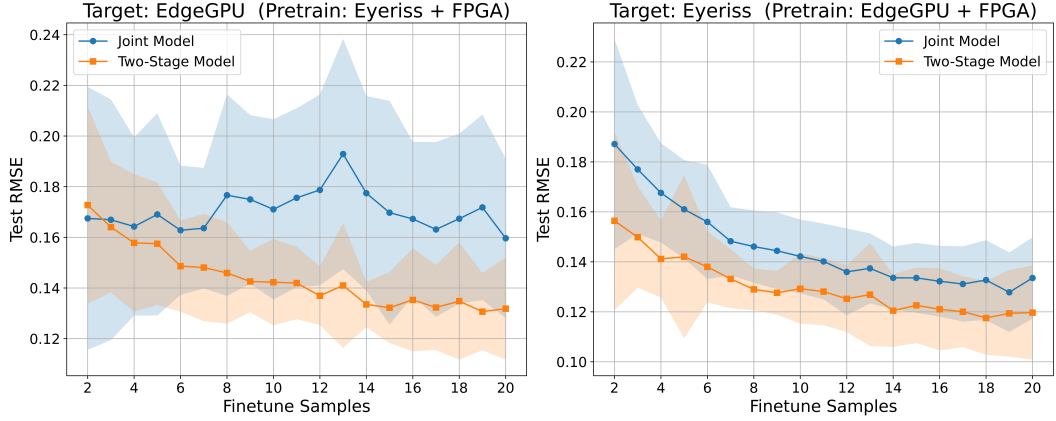}
  \caption{Two-Stage Energy Estimator vs. Joint Model under few-shot adaptation with 2--20 samples on HW-NAS-Bench~\cite{li_hw-nas-bench_2021}, using the NAS-Bench-201~\cite{dong_nas-bench-201_2019} search space and the EdgeGPU and Eyeriss hardware targets.}
  \label{fig:hwnasbench}
\end{figure}

\subsection{Energy Estimation with Sparse Hardware Samples}
\label{sec:more}

We next evaluate the proposed two-stage energy estimator under limited device-specific supervision. Figure~\ref{fig:hwnasbench} reports results on HW-NAS-Bench, using the NAS-Bench-201 search space and two hardware targets: EdgeGPU and Eyeriss. For each target device, we pretrain the estimator on the remaining hardware targets and then fine-tune on the target device using only 2 to 20 samples. We compare the proposed two-stage estimator against a joint model under the same training protocol and report test RMSE on the held-out target device \cite{li_hw-nas-bench_2021,dong_nas-bench-201_2019,guo_estimating_2025,zhang_ampere_2025}.

Figure~\ref{fig:hwnasbench} shows that the two-stage estimator achieves lower test RMSE than the joint model on both EdgeGPU and Eyeriss across nearly the full few-shot range. The improvement is visible even in the lowest-data regime and remains consistent as more target-device samples are added. These results indicate that separating generic architecture-level energy structure from hardware-specific residual correction yields more sample-efficient adaptation than directly fitting a single joint predictor \cite{guo_estimating_2025,zhang_ampere_2025}.

\begin{figure}[h]
    \centering
    \begin{subfigure}[b]{0.51\textwidth}
        \centering
        \includegraphics[width=\textwidth]{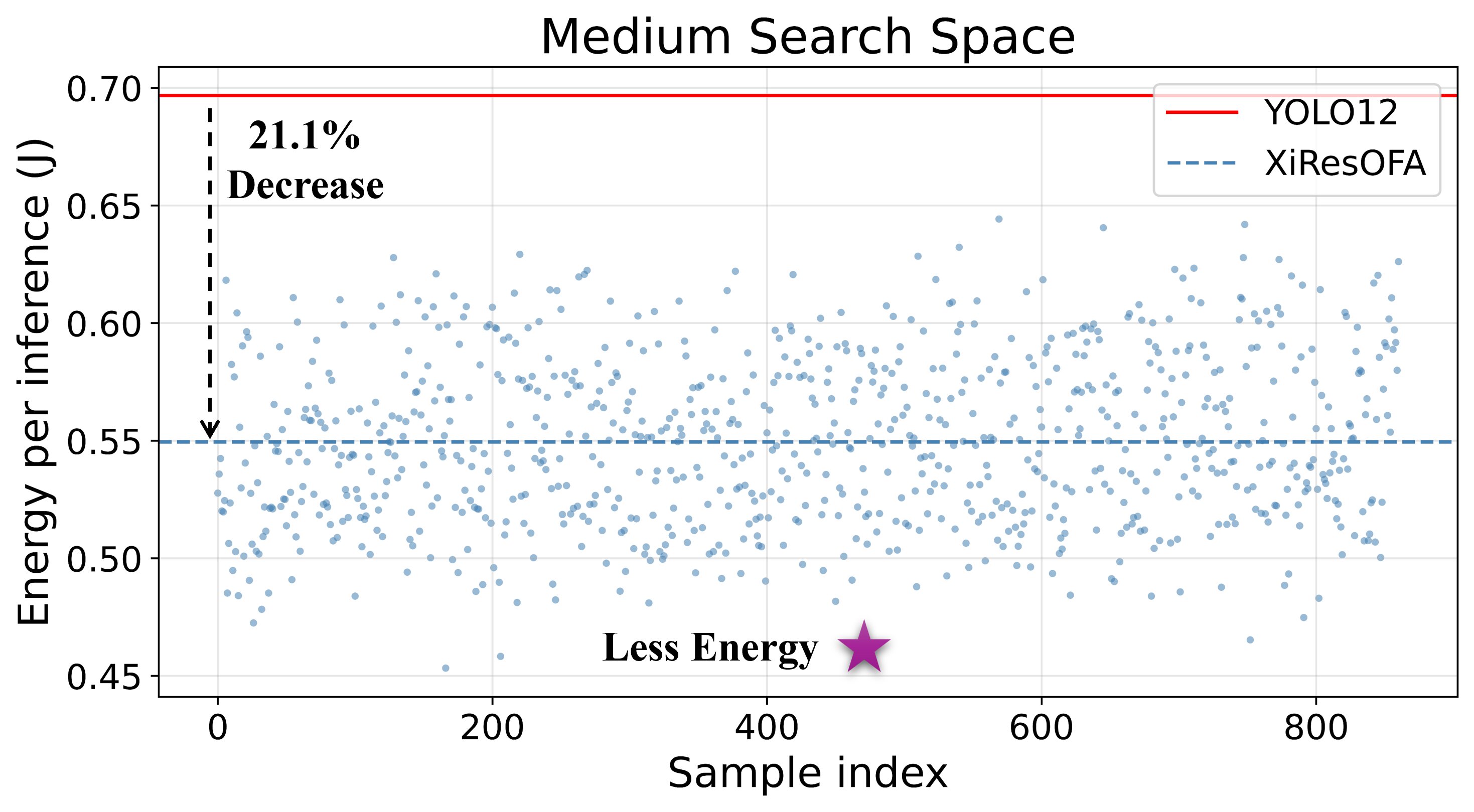}
        \caption{Energy distribution of randomly sampled XiResOFA medium models.}
        \label{fig:energy_space}
    \end{subfigure}
    \hfill
    \begin{subfigure}[b]{0.45\textwidth}
        \centering
        \includegraphics[width=\textwidth]{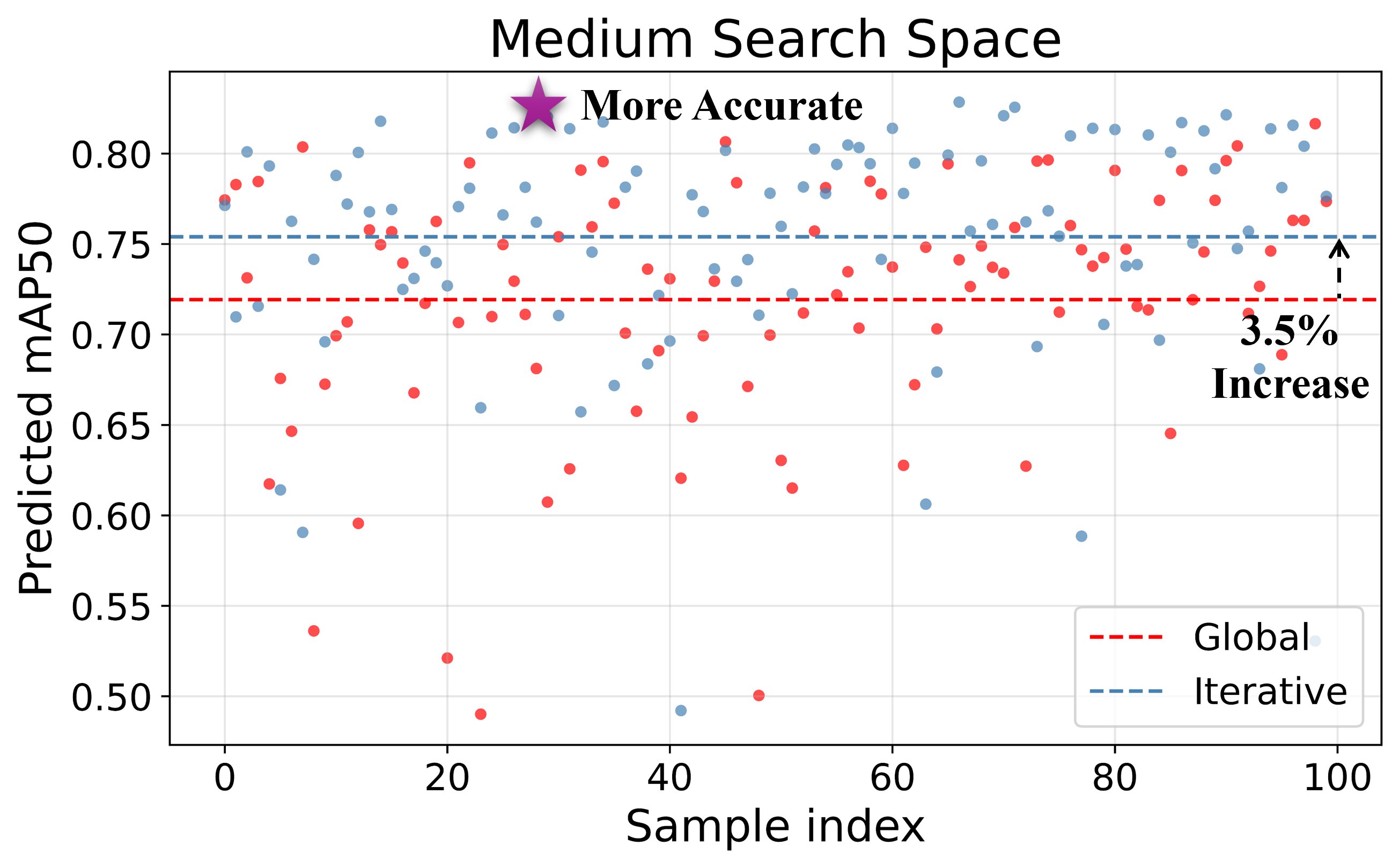}
        \caption{Predicted accuracy of sampled models from global and iterative search spaces.}
        \label{fig:acc_space}
    \end{subfigure}
    \caption{Energy and accuracy characteristics of the medium search space.}
    \label{fig:design_spaces}
\end{figure}

\subsection{Search Space Characteristics}

We next examine the properties of the proposed medium-scale search space in terms of both energy and accuracy. Figure~\ref{fig:design_spaces}(a) shows the energy distribution of randomly sampled XiResOFA models. Compared with the YOLOv12 medium baseline, the sampled models consistently occupy a lower-energy region, with the average energy reduced by \textbf{21.1\%}. This indicates that the proposed search space is biased toward more energy-efficient designs while still spanning a broad set of candidate architectures.

Figure~\ref{fig:design_spaces}(b) compares the predicted accuracy of sampled architectures from the global and iterative search spaces. The iterative search space yields a higher average predicted mAP50 than the global search space, with an improvement of about \textbf{3.5\%}. This suggests that the iterative decomposition produces a more favorable candidate region for optimization, concentrating search on architectures with stronger expected accuracy. Together, these results show that the proposed XiResOFA design space is both energy-efficient and structurally well suited for iterative search.

\section{Conclusion}
\label{sec:conclusion}

We presented an energy-adaptive framework for object detection on heterogeneous edge devices that combines an energy-aware search space, a two-stage energy estimator, and iterative neural architecture search. The proposed XiResOFA search space exposes architectural choices that directly control the accuracy--energy tradeoff, while the two-stage estimator enables energy-aware search under sparse hardware measurements. By searching for a single energy-efficient base architecture and scaling it into the XiYOLO family, our framework achieves strong accuracy--energy tradeoffs across GPU and NPU deployment. On PascalVOC, XiYOLO improves over YOLOv12 at the small and medium scales while substantially reducing energy on both devices. On COCO, XiYOLO again delivers large energy reductions relative to YOLOv12, especially at the small scale, while maintaining competitive medium-scale accuracy. We also show that XiYOLO achieves the best deployment-time energy behavior among compared medium-sized models and that the proposed two-stage estimator improves few-shot adaptation over a joint predictor on HW-NAS-Bench.

This work has both positive and negative societal implications. More energy-efficient edge perception can improve the practicality of search and rescue, disaster response, infrastructure inspection, and environmental monitoring by extending deployment time and reducing reliance on remote compute. At the same time, efficient onboard perception could also enable harmful uses such as persistent surveillance or reduced human oversight. Our study also has several limitations: the search is confined to a YOLOv12-derived detector family, the final model family is obtained through post-search scaling rather than direct multi-point search, and the evaluation focuses on selected GPU and NPU deployment settings. Despite these limitations, our results show that energy-aware architecture search can produce detector designs that are both accurate and energy-efficient for heterogeneous edge deployment.



\bibliographystyle{plainnat} 
\bibliography{biblio}







\newpage
\appendix

\section{Full Deployment Energy Results}

\begin{figure}[h]
  \centering
  \includegraphics[width=\textwidth]{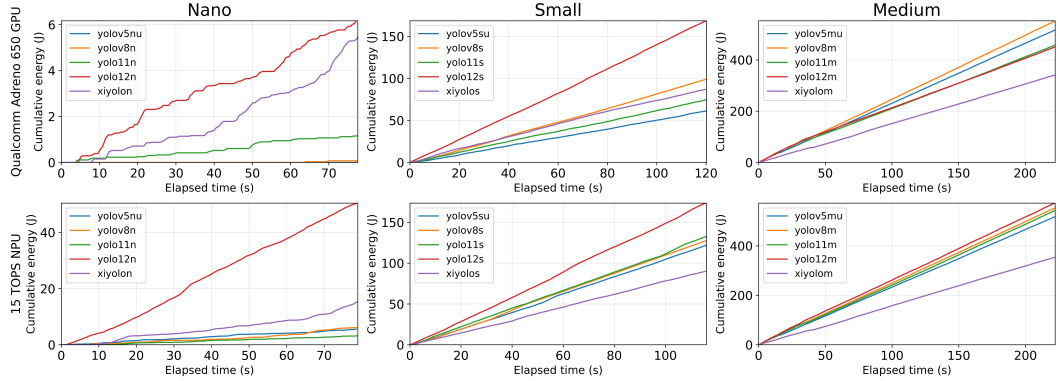}
  \caption{Full energy consumption vs. time benchmarks against other YOLO-series~\cite{jocher_yolov5_2020, jocher_yolov, tian_yolov12_2025} on the GPU (top) and NPU (bottom) for the nano (left), small (middle), and medium (right) model scales on the PascalVOC dataset.}
  \label{fig:fullenergy}
\end{figure}

Figure~\ref{fig:fullenergy} reports the full cumulative-energy trajectories for all nano, small, and medium models on the Qualcomm Adreno 650 GPU and the 15 TOPS NPU. These results complement the main-text medium-scale comparison by showing the complete deployment behavior across all scales and baselines.

Several trends are consistent across the full benchmark. First, the strongest energy advantages of XiYOLO appear at the medium scale. On the GPU, \textbf{XiYOLOm} achieves the lowest cumulative energy among all compared medium-sized models at \textbf{340.2 J}, compared with \textbf{449.5 J} for YOLOv12m, \textbf{456.5 J} for YOLO11m, \textbf{516.6 J} for YOLOv5mu, and \textbf{552.4 J} for YOLOv8m. On the NPU, \textbf{XiYOLOm} again outperforms all YOLO baselines at \textbf{352.9 J}, compared with \textbf{572.6 J} for YOLOv12m. These results are consistent with the main-text finding that the searched XiYOLO architecture provides the strongest deployment-time energy behavior in the medium regime.

Second, the relative gains at the small and nano scales are more mixed, but XiYOLO still remains competitive and often improves substantially over YOLOv12. On the GPU, \textbf{XiYOLOs} reduces cumulative energy relative to YOLOv12s from \textbf{168.6 J} to \textbf{86.9 J}. On the NPU, the same comparison drops from \textbf{173.5 J} for YOLOv12s to \textbf{90.0 J} for \textbf{XiYOLOs}. At the nano scale, XiYOLO also reduces cumulative energy relative to YOLOv12n on both accelerators, lowering GPU energy from \textbf{6.17 J} to \textbf{5.45 J} and NPU energy from \textbf{50.45 J} to \textbf{15.23 J}. However, some other baselines remain more energy-efficient at these lower-capacity operating points, especially YOLOv5 and YOLO11 in selected nano and small settings. Overall, the full results show that XiYOLO maintains the clearest advantage at the medium scale, while remaining competitive and often substantially more efficient than YOLOv12 across the broader deployment space.
    
\section{Additional Energy Estimation Results on ModalAI Sentinel}

\begin{figure}[t]
  \centering
  \includegraphics[width=\textwidth]{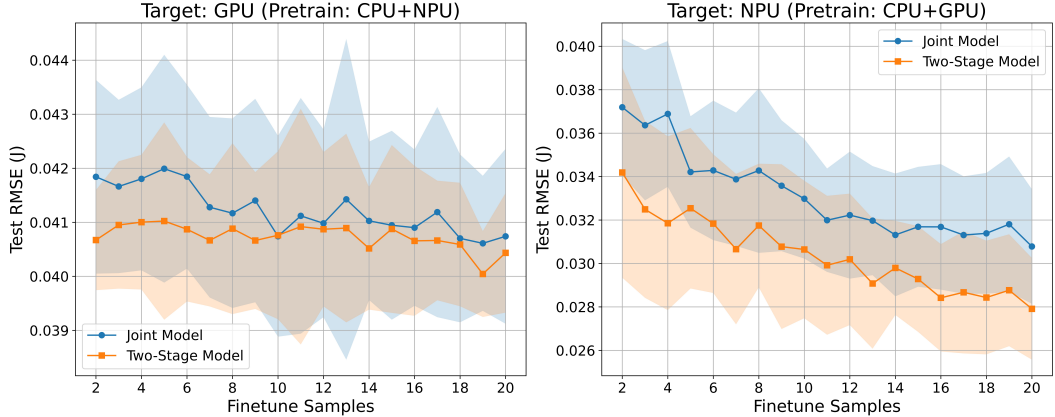}
  \caption{Two-Stage Energy Estimator vs. Joint Model on few data points from 2--20 samples on the ModalAI Sentinel Development Drone benchmark on CPU, GPU, and NPU using the proposed search space.}
  \label{fig:cpugpunpu}
\end{figure}

To further evaluate the proposed two-stage energy estimator on a real robotic edge platform, we include additional results on the ModalAI Sentinel Development Drone in Figure~\ref{fig:cpugpunpu}. In this experiment, we use 100 sampled architectures from the proposed search space benchmarked on the CPU, GPU, and NPU of the ModalAI Sentinel drone. For fairness, both the joint model and the two-stage model use the same generic architecture/device encoder. The architecture encoder one-hot encodes the XiResidualOFA design choices, while the device encoder one-hot encodes the target hardware. Energy values are normalized with respect to the range of each device. The joint model receives the concatenated architecture and device encoding and is pretrained on two devices before being fine-tuned on the target device. The proposed two-stage model first pretrains a generic predictor using only architecture encoding on the other two devices, then freezes this base predictor and fine-tunes a device-specific residual model on the target device using both architecture and device encoding. We evaluate few-shot adaptation from 2 to 20 fine-tuning samples, repeat each setting over 30 runs, and use the remaining target-device data as the test set \cite{guo_estimating_2025,zhang_ampere_2025}.

The results show that the two-stage model matches the joint model on CPU and outperforms it on GPU and NPU, with the clearest gains in the low-data regime. This supplementary experiment provides additional evidence that the proposed formulation remains effective on a real heterogeneous robotic platform, where accurate hardware energy measurements are especially limited.

\section{Architecture Ablation}

We perform an ablation study to isolate the effect of the two main design changes in the proposed block: replacing the default YOLOv12 bottleneck with XiNet-style convolution and adding a residual connection. Specifically, we compare three detector variants: the original YOLOv12n bottleneck design (\textbf{Base}), a variant with only XiNet convolution (\textbf{XiConv}), and the proposed residual XiNet-based block (\textbf{XiRes}) \cite{tian_yolov12_2025,ancilotto_xinet_2023,he_deep_2016}. Table~\ref{tab:block_ablation} reports results on PascalVOC using small models deployed on GPU.

The ablation shows that introducing XiNet convolution yields the largest energy reduction, decreasing energy from \textbf{21.3 J} to \textbf{11.3 J} per 30 inferences, with a small drop in detection performance. Adding the residual connection recovers and slightly improves accuracy, yielding the best \textbf{84.38 mAP50} and \textbf{65.79 mAP50-95} while preserving the low-energy behavior at \textbf{11.1 J}. These results indicate that XiNet convolution is primarily responsible for the efficiency gain, while the residual connection improves optimization and restores accuracy, together motivating the final XiRes design used in the search space.

\begin{table}[h]
  \centering
  \renewcommand{\arraystretch}{1.15}
  \caption{Ablation of the proposed block design. Base is the default YOLOv12~\cite{tian_yolov12_2025} bottleneck, XiConv adds XiNet convolution~\cite{ancilotto_xinet_2023}, and XiRes further adds a residual connection~\cite{he_deep_2016}. Results are measured on small models running on GPU on the PascalVOC dataset \cite{everingham_pascal_2010}.}
  \begin{tabular}{lccc}
  & \textbf{Base} & \textbf{XiConv} & \textbf{XiRes} \\
  \hline
  XiNet Convolution   &  & \checkmark & \checkmark \\
  Residual Connection &  &  & \checkmark \\
  \hline
  mAP50 (\%)          & 84.34 & 84.26 & \textbf{84.38} \\
  mAP50-95 (\%)       & 65.69 & 65.15 & \textbf{65.79} \\
  Energy / 30 Inf (J) & 21.3$\pm$ 0.93 & 11.3$\pm$ 1.1 & \textbf{11.1$\pm$ 0.8} \\
  \end{tabular}
  \label{tab:block_ablation}
\end{table}

\section{Additional Results on Compound Scaling and Average Power}

To further evaluate the effect of compound scaling, we compare the average power of scaled XiYOLO models against YOLO baselines at the small and medium scales. The goal of this experiment is to test whether scaling the searched base architecture can preserve strong detection accuracy while also reducing deployment power across different hardware platforms. Since higher-accuracy detectors generally require more computation, accuracy and energy usage are often positively correlated. This makes compound scaling a useful test of whether the searched architecture can maintain a favorable tradeoff as model capacity increases.

Table~\ref{tab:power} reports results on the COCO dataset for the Qualcomm Adreno 650 GPU, the 15 TOPS NPU, and the NVIDIA Jetson Orin Nano. At the small scale, XiYOLOs reduces average power relative to both YOLOv8s and YOLO12s on all three platforms, while remaining competitive in accuracy. At the medium scale, XiYOLOm again achieves the lowest average power across all three devices, while preserving accuracy close to the stronger YOLO baselines. These results show that compound scaling of the searched base architecture can effectively limit the loss in detection accuracy while still reducing deployment power, supporting the use of XiYOLO as a scalable detector family for heterogeneous edge devices.

\begin{table}[h]
  \centering
  \renewcommand{\arraystretch}{1.15}
  \caption{Average power comparison for scaled detector variants on COCO. Results are reported on the Qualcomm Adreno 650 GPU, 15 TOPS NPU, and NVIDIA Jetson Orin Nano. XiYOLO achieves the lowest average power at both the small and medium scales while maintaining competitive accuracy. \textbf{Bold} indicates best while \underline{underline} indicates second best.}
  \begin{tabular}{lcccc}
    \hline
    \multirow{2}{*}{\textbf{Method}} & \multirow{2}{*}{\textbf{mAP50-95}} & \multicolumn{3}{c}{\textbf{Average Power during Inference}} \\
    \cline{3-5}
     &  & \textbf{Adreno 650 (W)} & \textbf{15 TOPS NPU (W)} & \textbf{Orin Nano (W)} \\
    \hline
    YOLOv8s          & 44.9                   & \underline{0.82} & \underline{1.1} & 12.6          \\
    YOLO12s          & \textbf{48.0}          & 1.4              & 1.5             & \underline{12.4} \\
    \rowcolor{gray!15}
    \textbf{XiYOLOs} & \underline{46.5}                   & \textbf{0.73}    & \textbf{0.78}   & \textbf{10.9} \\
    \hline
    YOLOv8m          & 50.2                   & 2.5              & \underline{2.5} & \underline{13.5} \\
    YOLO12m          & \textbf{52.5}          & \underline{2.0}  & 2.6             & 13.8          \\
    \rowcolor{gray!15}
    \textbf{XiYOLOm} & \underline{50.7}                   & \textbf{1.5}     & \textbf{1.6}    & \textbf{12.6} \\
    \hline
  \end{tabular}
  \label{tab:power}
\end{table}

\section{Latency--Energy Tradeoff}

\begin{figure}[h]
  \centering
  \includegraphics[width=\textwidth]{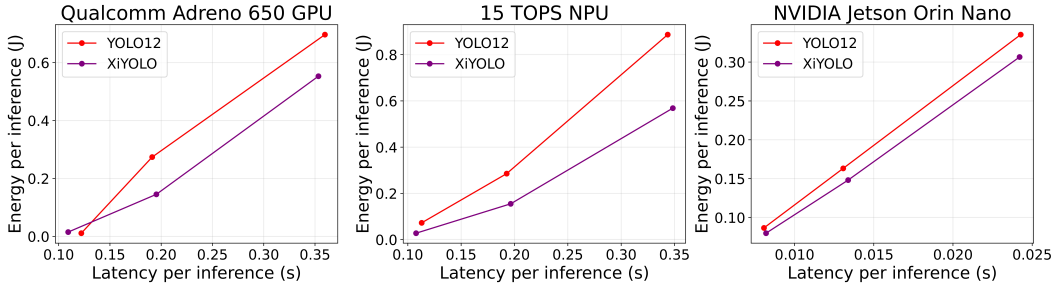}
  \caption{Latency--energy comparison between XiYOLO and YOLO12 on the Qualcomm Adreno 650 GPU, 15 TOPS NPU, and NVIDIA Jetson Orin Nano. XiYOLO consistently uses less energy at similar latency across all three platforms.}
  \label{fig:latency}
\end{figure}

Figure~\ref{fig:latency} compares XiYOLO and YOLO12 in the latency--energy plane on the Qualcomm Adreno 650 GPU, the 15 TOPS NPU, and the NVIDIA Jetson Orin Nano. Across all three devices, XiYOLO consistently achieves lower energy at similar latency. The gap is especially clear on the Adreno 650 GPU and 15 TOPS NPU, where XiYOLO remains well below YOLO12 across all operating points, while on the Jetson Orin Nano the improvement is smaller but still consistent. These results show that XiYOLO provides a better deployment tradeoff by reducing energy without introducing a meaningful latency penalty.

\end{document}